\documentclass[iicol]{sn-jnl}

\usepackage{graphicx}%
\usepackage{multirow}%
\usepackage{amsmath,amssymb,amsfonts}%
\usepackage{amsthm}%
\usepackage{mathrsfs}%
\usepackage[title]{appendix}%
\usepackage{xcolor}%
\usepackage{textcomp}%
\usepackage{manyfoot}%
\usepackage{booktabs}%
\usepackage{algorithm}%
\usepackage{algorithmicx}%
\usepackage{algpseudocode}%
\usepackage{listings}%
\usepackage{cite}
\usepackage[utf8]{inputenc}
\usepackage[numbers,sort,compress]{natbib}
\usepackage{tabularx}

\theoremstyle{thmstyleone}%

\theoremstyle{thmstyletwo}%

\theoremstyle{thmstylethree}%

\begin{document}

\title[Article Title]{Development and Validation of Heparin Dosing Policies Using an Offline Reinforcement Learning Algorithm}

\author[1]{\fnm{Yooseok} \sur{Lim}}\email{seook6853@soongsil.ac.kr}

\author[2]{\fnm{Inbeom} \sur{Park}}\email{inbeom@mju.ac.kr}

\author*[3]{\fnm{Sujee} \sur{Lee}}\email{sujeelee@skku.edu}

\affil[1]{\orgdiv{Department of Industrial and Information Systems Engineering}, \orgname{Soongsil University}, \orgaddress{\city{Seoul}, \postcode{06978}, \country{Republic of Korea}}}

\affil[2]{\orgdiv{Department of Industrial and Management Engineering}, \orgname{Myongji University}, \orgaddress{\city{Yongin}, \postcode{17058}, \country{Republic of Korea}}}

\affil*[3]{\orgdiv{Department of Systems Management Engineering}, \orgname{Sungkyunkwan University}, \orgaddress{\city{Suwon}, \postcode{16419}, \country{Republic of Korea}}}

\abstract{Appropriate medication dosages in the intensive care unit (ICU) are critical for patient survival. Heparin, used to treat thrombosis and inhibit blood clotting in the ICU, requires careful administration due to its complexity and sensitivity to various factors, including patient clinical characteristics, underlying medical conditions, and potential drug interactions. Incorrect dosing can lead to severe complications such as strokes or excessive bleeding. To address these challenges, this study proposes a reinforcement learning (RL)-based personalized optimal heparin dosing policy that guides dosing decisions reliably within the therapeutic range based on individual patient conditions.

A batch-constrained policy was implemented to minimize out-of-distribution errors in an offline RL environment and effectively integrate RL with existing clinician policies. The policy's effectiveness was evaluated using weighted importance sampling, an off-policy evaluation method, and the relationship between state representations and Q-values was explored using t-SNE. Both quantitative and qualitative analyses were conducted using the Medical Information Mart for Intensive Care III (MIMIC-III) database, demonstrating the efficacy of the proposed RL-based medication policy. Leveraging advanced machine learning techniques and extensive clinical data, this research enhances heparin administration practices and establishes a precedent for the development of sophisticated decision-support tools in medicine.}

\keywords{Reinforcement Learning (RL),
Personalized Heparin Dosing Policy, 
Batch-Constrained Policy, 
Weighted Importance Sampling, 
Medical Information Mart for Intensive Care III (MIMIC-III)}

\maketitle

\section*{Highlights}\label{highlights}
\begin{itemize}
    \item Developed a batch-constrained reinforcement learning (RL) approach using electronic medical record (EMR) data to optimize heparin dosing in intensive care units (ICUs).
    \item Integrated an expert behavior network to mitigate Q-value overestimation, achieving a balanced estimation between behavioral and optimal policies.
    \item Demonstrated superior performance of the Batch-Constrained Q-learning (BCQ) algorithm compared to traditional deep RL approaches.
    \item Applied t-SNE analysis to validate the RL policy's accurate learning of reward function objectives, highlighting high-value states conducive to achieving desired outcomes.
    \item Emphasized the potential of AI-driven decision support systems in enhancing thrombosis treatment strategies for clinicians and healthcare managers.
\end{itemize}

\section{Introduction}\label{sec1}

Thrombosis, characterized by blood clot accumulation in a vessel, disrupts circulation due to blockages. The symptoms of this condition can vary depending on the location and size of the clot, including swelling, pain, and shortness of breath. 
Untreated thrombosis can lead to blockage of the pulmonary artery, which can cause severe symptoms such as chest pain and difficulty in breathing, potentially resulting in death. Treating such clots typically involves administering anticoagulants such as heparin to prevent further clotting or thrombolytic agents to dissolve existing clots. 
However, anticoagulants are susceptible drugs whose effects are not immediately measurable; therefore, continuous monitoring is essential for guidance of an appropriate therapeutic range. 
Inappropriate dosing can have significant consequences, including strokes and excessive bleeding, with about 7,000 related deaths annually in the United States alone \cite{1}.

Traditionally, clinicians determine heparin doses using basic protocols, personal experience, and ongoing patient monitoring, methods that may not adequately account for individual patient variability \cite{2}.
This complexity necessitates reliable therapeutic tools to aid physicians in making precise dosing decisions. To this end, this study introduces a reinforcement learning (RL)-based, personalized heparin dosing policy aimed at optimally guiding therapy within the therapeutic range according to individual patient conditions. Heparin administration is framed as a Markov decision process (MDP), where clinicians assess the patient's condition and blood clotting status, administer medication, and adjust based on the activated partial thromboplastin time (aPTT) test results. This test, which simulates blood clotting, is crucial for monitoring and adjusting treatment until the desired clotting efficacy is achieved.

Previous approaches to heparin dosing include the use of partially observable MDPs (POMDPs), leveraging models like the hidden Markov model (HMM) and Q-learning to develop personalized protocols \cite{3,4}. 
However, typical off-policy deep reinforcement learning (DRL) algorithms struggle with extrapolation errors when training data deviates from actual distributions, often overestimating the value of new state-action pairs \cite{5}. This overestimation can lead to significant learning challenges in offline settings where direct interaction with the environment is not feasible. To address these limitations, we employed a batch-constrained deep Q-learning (BCQ) \cite{6} algorithm designed to minimize extrapolation errors and improve the accuracy of Q-value estimations.

Previous studies have typically evaluated RL policies using trajectory comparisons and statistical tests \cite{2,3,4,5}, which may not accurately determine if the RL has optimized rewards effectively, particularly in dynamic decision-making scenarios. In offline RL, where direct environmental interaction is absent, traditional evaluation methods struggle to assess the true efficacy of a policy, as they cannot simulate real-time feedback or explore the effects of policy adjustments. To address these challenges, this study employs weighted importance sampling (WIS) \cite{9} for policy evaluation, an off-policy method crucial for assessing the agent's learning effectiveness in the absence of direct environmental interaction.

Furthermore, to enhance our understanding of the learned RL policies, this study utilizes  t-distributed Stochastic Neighbor Embedding (t-SNE), a dimensionality reduction technique that excels in visualizing high-dimensional data. By applying t-SNE, we can visually inspect the clustering of state representations, providing insights into the agent’s decision-making processes. This visualization helps demonstrate how different patient states are managed under the RL policy, offering a clear view of the policy’s effectiveness and its alignment with clinical outcomes.

In summary, this study leverages a batch-constrained approach to reduce extrapolation errors in off-policy algorithms, employs the WIS technique for quantitative performance evaluation of the agent, and utilizes t-SNE to verify if the RL policy accurately learned the objectives set by the designed reward function. Our approach involves:

\begin{itemize}
\item Implementing the BCQ method to harness clinician-implied behavior in the data, mitigating extrapolation errors and enhancing the stability of learning.
\item Quantitatively assessing the learning performance using WIS, a robust off-policy evaluation technique.
\item Analyzing the policy derived by the RL agent with t-SNE to evaluate its alignment with, and differences from, clinician practices.
\end{itemize}

The organization of this paper is as follows: Section \ref{sec2} reviews relevant literature on RL and off-policy evaluation methods. Section \ref{sec3} details the materials and methods used in this study. Section \ref{sec4} presents and discusses the experimental results. Finally, Section \ref{sec5} concludes the study.

\section{Related Work}\label{sec2}

\subsection{Heparin Treatment}
Healthcare studies aiming to support clinician decision-making by introducing artificial intelligence systems have been conducted in various fields such as disease detection \cite{10,11}, health management \cite{12,13}, and drug administration \cite{14,15}. In particular, studies using supervised learning and RL have been proposed for optimizing heparin drug administration, aiming to control the degree of blood clotting in critically ill patients to maintain a normal state. Ghassemi et al. \cite{16} proposed an optimal initial-dosing prediction model using logistic regression. The model demonstrated that AI tools and electronic medical record (EMR) data can be utilized to quickly bring blood coagulation levels in critically ill patients within the therapeutic range. However, the decision support tools in these approaches only consider the static state of the moment to suggest the initial dose and do not consider the continuous nature of the patient care process, which requires ongoing care. Most medical practices, not only medication administration, are continuous processes in which patients interact with physicians to achieve optimal health. Consequently, these approaches have fundamental limitations.

RL solves continuous decision-making problems and is ideally suitable for heparin-dosing problems. Nemati et al. \cite{3} proposed an RL algorithm to determine the optimal dose of unfractionated heparin (UH) for a patient and to reduce the risk of health deterioration owing to medication errors. It defines the heparin administration process as a problem with a continuous state and discrete actions considering a partially observable Markov decision process (POMDP), and combines a discriminative hidden Markov model (DHMM) and a Q-learning model to solve drug administration problems. Baucum et al. \cite{4} proposed a method for optimizing the heparin dosing problem from a model-based RL perspective, which combines a supervised learning-based model with an RL algorithm to guide an agent to learn reliably. The aforementioned study utilized a transition variational autoencoder (tVAE) as an environmental model that inputs the patient state and clinician action to generate the subsequent state. As an RL algorithm, an asynchronous advantage actor–critic (A3C) was used, which has both a policy and value network, to optimize the heparin dosing policy. This system maximizes the effectiveness of RL by building and utilizing an environmental model rather than directly using the given data for learning.

\begin{figure*}[t]
    \centering
    \includegraphics[width=0.9\linewidth]{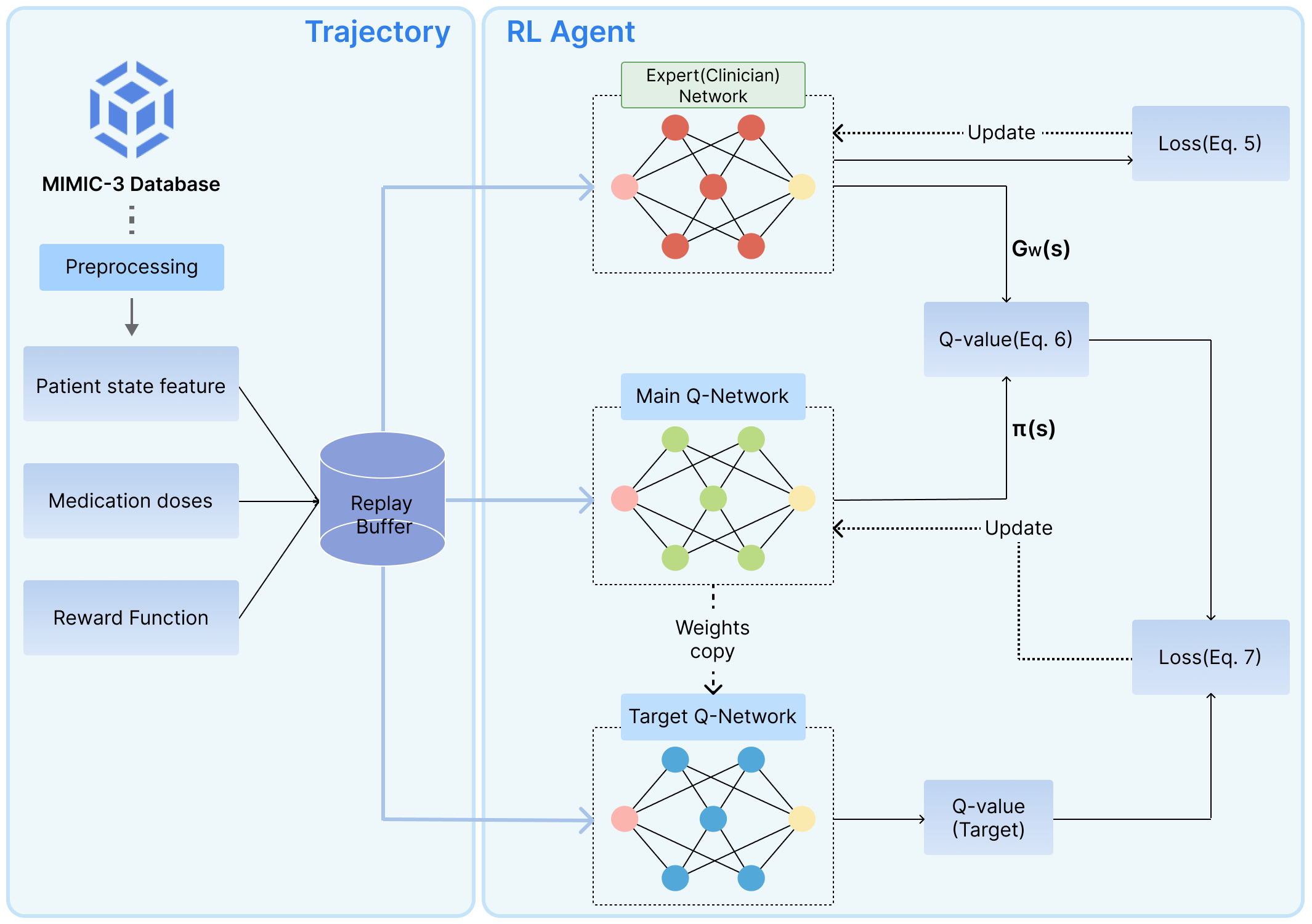}
    \caption{Diagram of the Proposed Reinforcement Learning Framework}
    \label{fig: figure1}
\end{figure*}

\subsection{Offline RL}
Deep RL has recently demonstrated promising results in specific fields, such as games \cite{17,18,19} and robot control \cite{20}. However, unlike the supervised learning methods that have been utilized in various fields based on large amounts of data, there are several conditions stemming from the MDP and Bellman Equation for applying RL. First, the problem must be defined as a sequential process that satisfies the MDP over time. Second, an active interaction with the environment is essential to evaluate and improve the performance of the policy during the learning process. The second constraint is a major factor that complicates the application of RL to real-world decision-making problems. Interacting with the real world (driving a car or treating a patient) is costly and risky. Offline RL \cite{5,6,21,22} is a methodology designed to overcome these issues and extend the generality of RL; it learns an optimal policy using only previously collected data without interacting with the environment. This approach is widely used in robotics \cite{robot1,robot2,robot3}, autonomous driving \cite{driving1,driving2,driving3}, recommender systems \cite{recommender1,recommender2}, and language dialogue \cite{language1,language2}.

Commonly used off-policy algorithms \cite{17,23} suffer from learning difficulties in offline settings due to two significant problems: bootstrapping and overfitting with out-of-distribution behaviors \cite{5}. These issues cause algorithms to either overestimate or incorrectly estimate the value of the state-action pairs that differ from the batch distribution. Since RL involves sequential processes, the effects of overestimation are cumulative, leading to learning failure. In online learning, agents can interact with the environment and validate their estimated values, which is not possible in offline learning.

To address these issues, two primary strategies are commonly employed: 1) enhancing the network architecture \cite{5,6}, and 2) incorporating constraints into the loss function \cite{21,22}. Fujimoto et al. \cite{5} proposed the BCQ method to augment the existing deep Q-network (DQN) framework. BCQ integrates a Q-network with a state-conditioned generative model, thereby restricting the agent's actions to those previously observed in the batch data. This approach guides the agent to select actions that not only resemble those in the batch data but also possess the highest value, thereby facilitating effective learning without interaction with the environment. In this study, we implemented the BCQ algorithm to effectively address the outlined challenges.

\subsection{Off Policy Evaluation}
In RL, the cumulative reward of a policy is typically estimated to assess its performance. In environments where interaction is possible, the learned policy can directly measure returns through such interactions. However, in offline RL, where no interaction with the environment is possible, alternative evaluation methods are required. Importance sampling serves as a robust technique for evaluating a policy using trajectories collected from a different policy. Naïve importance sampling allows for the computation of unbiased estimators from batch trajectories but suffers from high variance as trajectory length increases due to the product of importance-sampling weights. Weighted importance sampling (WIS) \cite{wis}, which incorporates a self-normalizing term, significantly reduces the variance of naive estimates. In practice, WIS provides consistent estimates with markedly lower variance.

\begin{table*}[h!]
\caption{Summary of the feature statistics}\label{tab:table1}
\centering
\resizebox{0.8\textwidth}{!}{%
\renewcommand{\arraystretch}{1.2}
\begin{tabular*}{\textwidth}{@{\extracolsep\fill}lllll@{\extracolsep\fill}}
\toprule[1.5pt]
Category     & Features& Mean (SD)  & Features    & Mean (SD)   \\ \midrule\hline
Demographics &  Admissions (N)& 1911& Male (\%)& 58\\
& Age& 65.9 (15.5)  & Weight& 89.4 (27.9)\\
Vital Signs  & DBP& 58.2 (14.7)  & GCS& 13.7 (2.9)    \\
& SBP& 119.1 (21.4) & Temperature & 36.9 (1.6)    \\
& RR& 19.6 (5.1)   &&     \\
Lab values& aPTT& 67.3 (34.4)   & WBC& 11.2 (6.6)    \\
& HGB& 10.6 (1.9)   & Platelets   & 226.4 (113.8) \\
& PT& 15.7 (4.2)   & ACD& 40.4 (8.9)     \\
& Creatinine& 1.6 (3.7)& Bilirubin   & 10 (1.8)      \\
& INR& 1.4 (0.5)    &&  \\ 
\bottomrule[1.5pt]
\end{tabular*}%
}
\end{table*}

\section{Materials and Methods}\label{sec3}

Fig. \ref{fig: figure1} presents a high-level diagram of the proposed approach. In clinical settings, heparin dosing protocols are defined based on the body weight, and the effect of treatment is measured using the aPTT laboratory test \cite{16}, which assesses the blood clotting ability. Since the effects of treatment are not immediate, the aPTT test is typically performed 4 to 6 hours after administering heparin. Based on the test results and other clinical indicators, clinicians adjust the heparin dose until clotting levels normalize.

The proposed approach uses RL to formulate an optimal dosing strategy that maintains patient blood coagulation levels within a therapeutic range. For training the RL model, we utilize a trajectory $\tau$ comprising data from $N$ patients. Each patient's data includes laboratory test results $s$, clinician-administered actions $a$, and rewards $r$ defined by the aPTT levels.

\begin{equation} \label{eq1}
\begin{split}
    \tau = [\{{(s^{(1)}_{i,1},a^{(1)}_{1,1},r^{(1)}_{1,1}),...,(s^{(1)}_{i,t_1},a^{(1)}_{1,t_1},r^{(1)}_{1,t_1})}\},...,\\
    \{(s^{(N)}_{i,1},a^{(N)}_{1,1},r^{(N)}_{1,1}),...,(s^{(N)}_{i,t_n},a^{(N)}_{1,t_n},r^{(N)}_{1,t_n})\}]
\end{split}
\end{equation}

Each trajectory varies in length, corresponding to the specific duration of treatment for each patient. The state $s^{(n)}_{i,t_n}$ represents $i$-dimensional laboratory values for the $n$th patient at time $t_n$, while $a^{(n)}_{1,t_n}$ and $r^{(n)}_{1,t_n}$ include $1$-dimensional actions and rewards, respectively. Missing values in $r^{(n)}$ are possible due to the timing of aPTT measurements post-heparin administration. The goal is to learn an optimal RL policy $\pi^*$ that maximizes the cumulative reward based on the trajectories $\tau$.

\begin{figure*}[h!]
    \centering
    \includegraphics[width=0.65\textwidth]{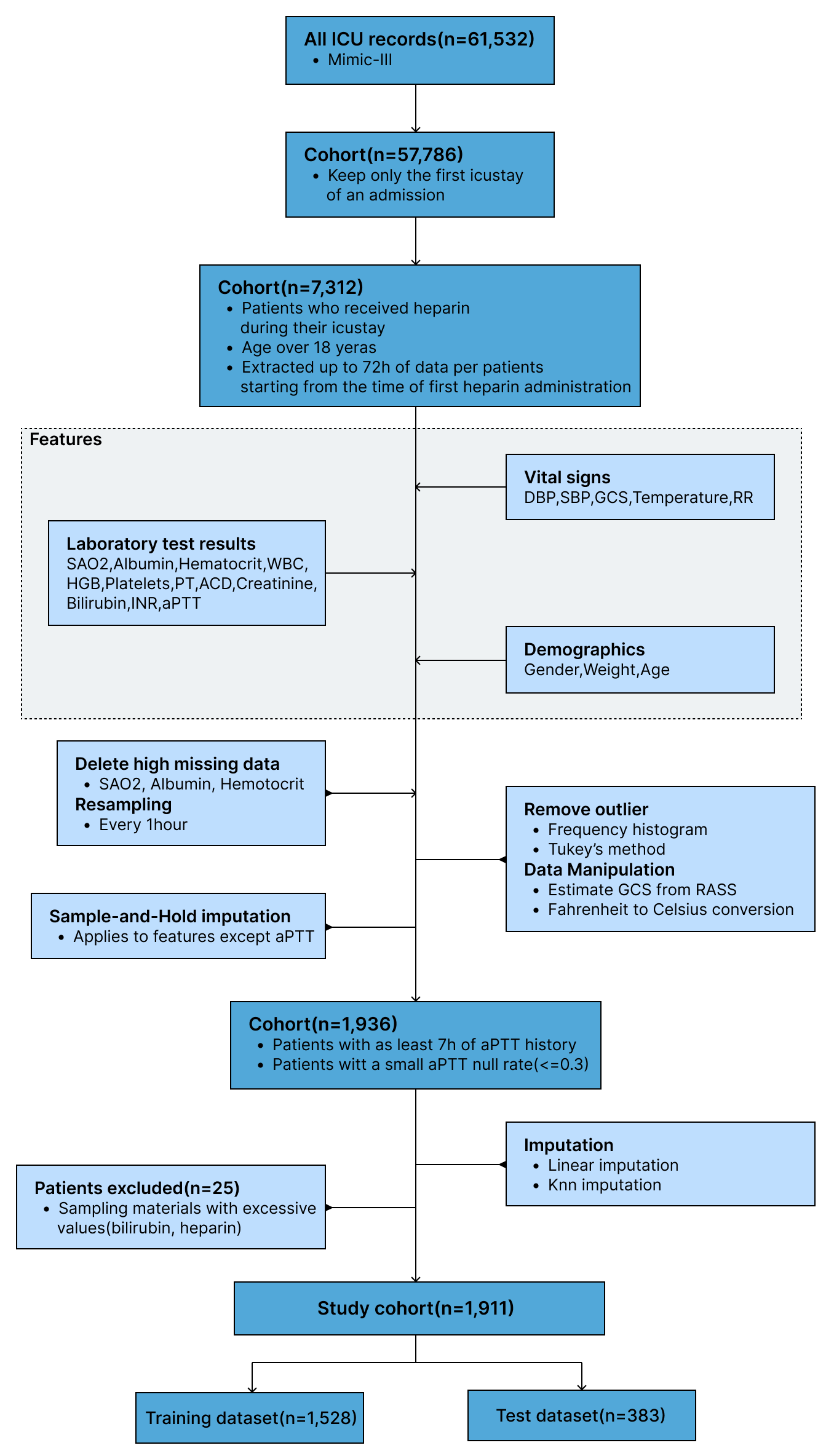}
    \caption{Flowchart of Data Extraction and Preprocessing}
    \label{fig: preprocessing}
\end{figure*}

\subsection{Dataset}
MIMIC-III data \cite{8} was utilized, which is a de-identified health-related public database of more than 40,000 patients admitted to the Beth Israel Deaconess Medical Center ICU between 2001 and 2012. It contains various types of information regarding patient care, such as the demographics, laboratory test results, procedures, vital sign measurements, and caregiver notes.

To obtain the data of patients receiving intravenous heparin in the ICU, we extracted approximately 21 laboratory test results, vital signs, and heparin dosing results related to the health and anticoagulation of patients aged $> 18$ years. 
Since the MIMIC-III database compiled this information from two distinct sources, we standardized the measurement units to `units' for heparin doses, aligning disparate records that originally included `units', `ml', and `units/ml'. We then collected data from each patient ranging from a minimum of 7 to a maximum of 72 hours following the administration of the initial heparin dose.

Given that RL typically requires data with a sequential flow over fixed time intervals, we resampled the data hourly. During this resampling, we averaged multiple occurrences within the same hour. However, as laboratory tests and vital sign checks in hospitals are performed based on clinical need rather than at fixed intervals, this resampling increased the incidence of missing values. We addressed this by calculating the missing rates for each feature and excluding those with high missing rates. Remaining missing values, post-outlier removal, were imputed using the Sample-and-Hold and k-nearest neighbor (KNN) methods \cite{3}.

For the RL model, we utilized the following variables: age, gender, Glasgow coma score (GCS), diastolic and systolic blood pressure (DBP and SBP), respiratory rate (RR), hemoglobin (HGB), temperature, white blood cell count (WBC), platelet count, activated partial thromboplastin time (aPTT), prothrombin time (PT), arterial carbon dioxide (ACD), creatinine, bilirubin, international normalized ratio of prothrombin (INR), and weight. We applied z-score normalization to ensure stable training. The dataset comprises 17 features, corresponding to 1,911 hospitalization records and 1,838 unique patients. Detailed statistics of the features are presented in Table \ref{tab:table1} and the complete flow of data extraction, preprocessing, and normalization processes described above is illustrated in Fig. \ref{fig: preprocessing}.

\subsection{State Space and Action Space}
\begin{figure}[h!]
    \centering
    \includegraphics[width=1.0\linewidth]{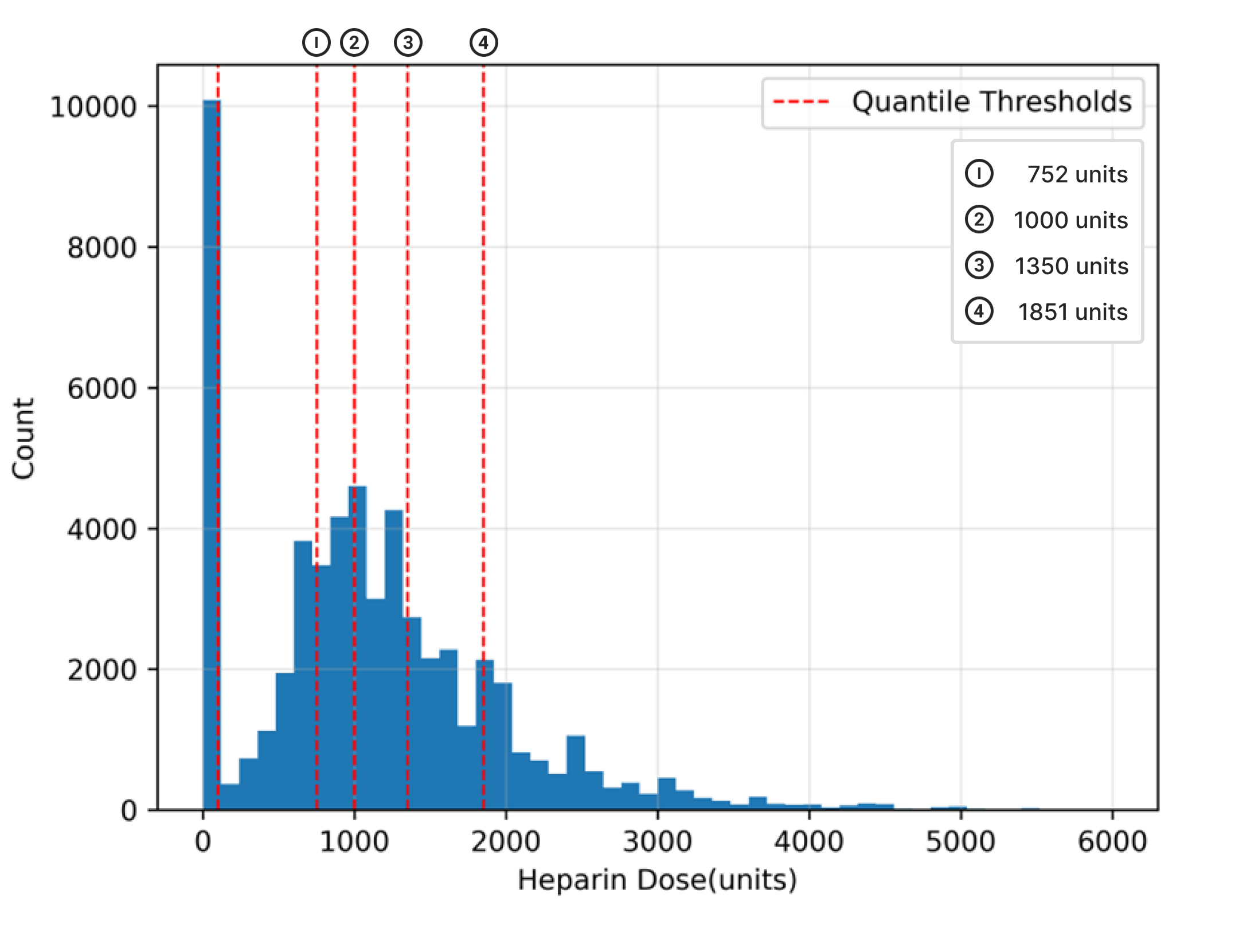}
    \caption{Distribution and Categorization of Heparin Doses}
    \label{fig: figure2}
\end{figure}

The state space is defined by the characteristics that change in response to actions and rewards. From the 17 features extracted from the MIMIC-III database, excluding aPTT, the remaining 16 features reflect patient health and anticoagulation levels, making them suitable for representing the state of a patient. These features effectively capture the dynamics related to changes in heparin medication and aPTT levels.
Heparin dosing was discretized into six categories for the action space. A category zero was designated for cases with no heparin administered, and the remaining dosing spectrum was divided into five categories defined by the 20th, 40th, 60th, and 80th percentiles \cite{3}. Fig. \ref{fig: figure2} illustrates the distribution of heparin doses and the corresponding categorization thresholds, demonstrating a total of six dosing categories represented as $a_t \in \{0,1,2,3,4,5\}$.

\subsection{Reward Formula}
\begin{figure}[h!]
    \centering
    \includegraphics[width=1.0\linewidth]{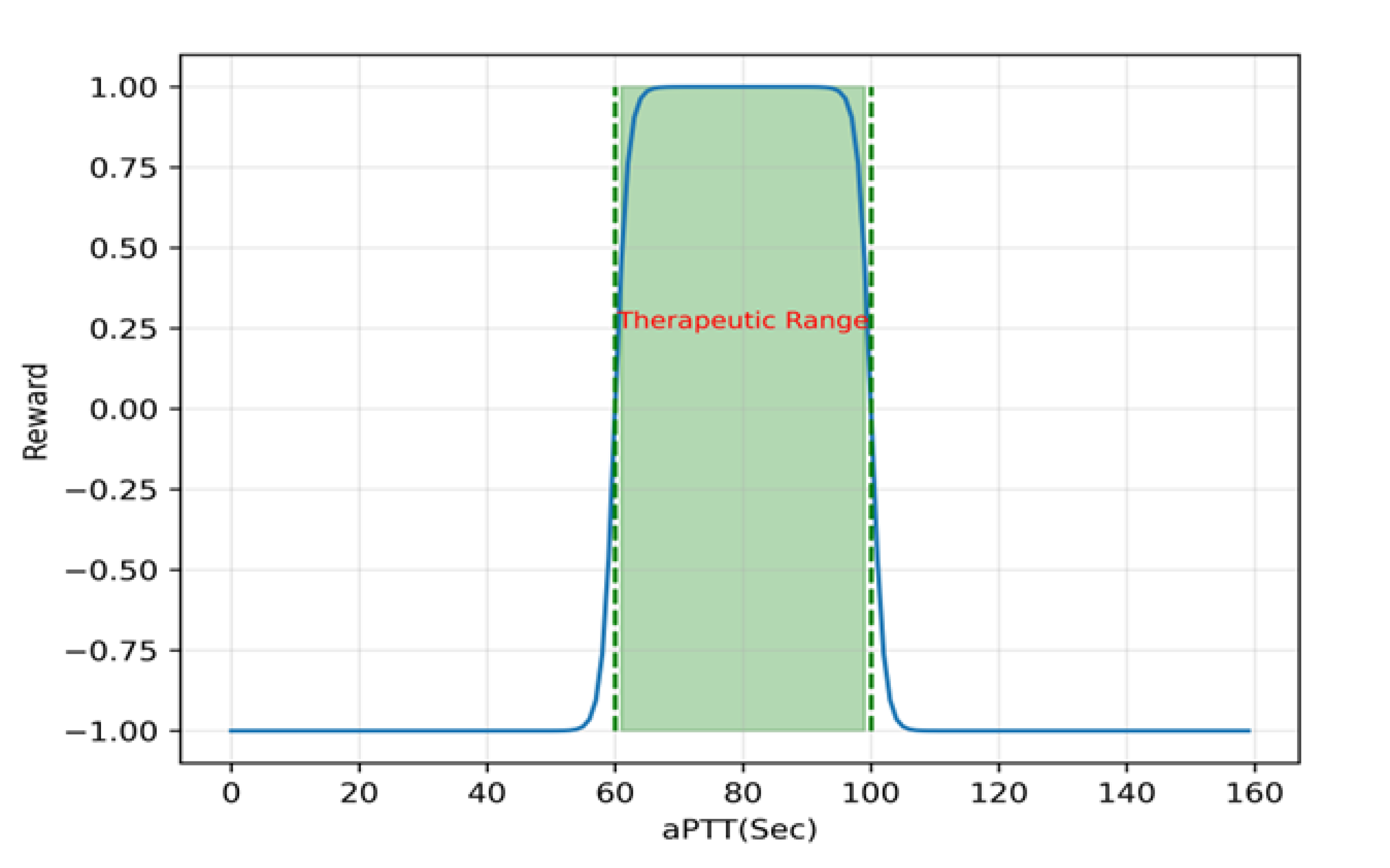}
    \caption{Reward Allocation Based on aPTT Levels}
    \label{fig: figure3}
\end{figure}

A fundamental component of RL is the reward function, which motivates the agent to achieve the problem's objective. This function either rewards or penalizes the agent's behaviors, depending on their outcomes. An agent acts in state $S_t$ to reach the next state $S_{t+1}$ and receives a reward $r$. In our study, the primary goal is to regulate the patient's blood clotting levels within a therapeutic range. The aPTT is a critical indicator of blood clotting ability and thus serves as the basis for our reward metric. Utilizing aPTT, Nemati et al. \cite{3} proposed a reward function, which is defined in (\ref{eq2}), that has been widely adopted in subsequent studies \cite{4,7}. This function allocates a reward of approximately $1$ for aPTT values within the desired therapeutic range of 60-100 seconds, and a reward of approximately $-1$ for values outside this range, as illustrated in Fig. \ref{fig: figure3}.

\begin{equation} \label{eq2}
    R_t = {2\over1+e^{-{(aPTT-60)}}} - {2\over1+e^{-{(aPTT-100)}}} -1
\end{equation}

\subsection{Batch-Constrained Policy}
To enhance learning efficiency in an offline environment and prevent Q-function overestimation while incorporating clinician drug administration data, we employed the BCQ algorithm, which combines Double DQN and $G_w$.
The objective of the Double DQN is to maximize the expected return, and the state and action value functions are given by (\ref{eq3}) and (\ref{eq4}) where $\theta^{'}$ denotes the target network, $\theta$ refers to the current network, $r$ is the reward, and $\gamma$ is the decay rate. The value of the next state has a recursive relationship with that of the current state.

\begin{eqnarray} \label{eq3}
    &V(s) = E[r+\gamma V(s')] \\\label{eq4}
    &Q(s,a) = E[r + \gamma Q_{\theta^{'}}(s', argmax_{a^{'}}Q_\theta(s',a'))]
\end{eqnarray}

We used $G_w$ to leverage the clinician dosing knowledge for RL. The value network is trained to estimate the optimal value, which depends on the reward defined using the aPTT. However, in real-world medical environments, clinician decisions do not always positively correlate with aPTT results due to various circumstances. To ensure safety beyond merely maximizing rewards, we incorporated clinician dosing policies into the learning process to develop safe patient treatment strategies. $G_w$ helps the RL agent learn a policy that maximizes the reward within a range of actions that satisfy the clinician dosing knowledge. The loss is expressed by (\ref{eq5}).

\begin{equation} \label{eq5}
    L(\omega) = -\sum^{m}_{i=1} y_i \log(\hat{y_i}), \; \hat{y}_i \sim G_\omega(s),\; y_i \sim a_{\pi_b}
\end{equation}

In (\ref{eq5}), $\pi_b$ denotes the behavioral policy. The RL agent adopts the expert policy in the learning process, excluding actions that do not align with the $G_w$ policy. The agent policy is given by (\ref{eq6}).

\begin{equation} \label{eq6}
    \pi(s) = argmax_{a|{{G_\omega(a|s)}\over \max_{\hat{a}}G_{\omega}(\hat{a}|s)}>\tau}Q_{\theta}(s,a)
\end{equation}

In the RL policy, actions falling below the threshold $\tau$ are disregarded. We scale the actions based on their maximum probability within $G_w$, allowing only those actions whose relative probability exceeds a specified threshold. The loss is expressed as follows:

\begin{equation} \label{eq7}
\begin{split}
    L(\theta) = E[(r + \gamma max_{a'|{{G_\omega(a'|s')}\over \max_{\hat{a}}G_{\omega}(\hat{a}|s')}>\tau}Q_{\theta}(s',a')) \\-Q_{\theta}(s,a)]
    \end{split}
\end{equation}

Methods that reflect the clinician dosing knowledge in the policy also have advantages in terms of value-learning. In an offline RL environment, a typical off-policy algorithm may overestimate a previously unseen $(s, a, r, s')$ tuple while learning the Q-value. Consequently, overestimation hinders the estimation of accurate Q-values and causes the model to fail to learn. In BCQ, $G_w$ limits the computation of Q-functions for state-action pairs that the RL agent did not observe during learning, thereby reducing the likelihood of overestimation and enabling stable learning.

\subsection{Off-Policy Evaluation with WIS}
WIS was employed to quantitatively assess the policy. The evaluation uses importance sampling (IS) with data derived from behavioral policies to estimate the expected value of the target policy. It computes the importance ratio for each state-action pair in the dataset, which is then used to estimate the expected return of the target policy. The importance ratio is defined as the probability of a sample occurring in the target distribution divided by its probability in the behavioral distribution. The IS calculation is detailed in (\ref{eq8}) and involves adjusting for the variance inherent in this method.

\begin{equation} \label{eq8}
\begin{split}
    J(\pi_\theta) &= E_{{\tau} \sim {\pi_{b}(\tau)}}[\frac{\pi_{\theta}(\tau)}{\pi_{b}(\tau)} \sum^{H}_{t=0} \gamma^{t} r(s,a)]\\
                &= E_{{\tau} \sim {\pi_{b}(\tau)}}[\prod^{H}_{t=0} \frac{\pi_{\theta}(\tau)}{\pi_{b}(\tau)} \sum^{H}_{t=0} \gamma^{t} r(s,a)]\\
                &\approx \sum^n_{i=1} w^i_H \sum^H_{t=0} \gamma^t r^i_t,
\end{split}
\end{equation}
where $H$ is the time horizon of the sample, $w^i_t = {1 \over n} \prod^t_{t'=0}{\pi_\theta(a^i_{t'}|s^i_{t'}) \over \pi_b(a^i_{t'}|s^i_{t'})}$, and $\{s^i_0, a^i_0, r^i_0, s^i_1,...\}^n_{i=1}$ is $n$ trajectory samples of $\pi_b(\tau)$.

While IS provides a straightforward way to estimate policy values, its high variance can make accurate estimations challenging. To address this, WIS was utilized as described in (\ref{eq9}), providing a more stable estimate by self-normalizing the importance ratios.

\begin{equation} \label{eq9}
    J(\pi_\theta) \approx \frac{\sum^n_{i=1} w^i_H \sum^H_{t=0} \gamma^t r^i_t}{\sum^n_{i=1} w^i_H}
\end{equation}

\section{Experimental Results}\label{sec4}

\subsection{Network Architecture and Hyper-parameters}
Unless specified otherwise, all networks share the same architecture and hyper-parameters. The architectures of both the Q-network and the clinician network are illustrated in Fig. \ref{fig: figure1}.

The state inputs are processed through a three-layer linear network. The input layer has a dimension of 16, each linear layer contains 256 nodes, and the output dimension is 6. The Q-network outputs the Q-values for each action, with ReLU activation functions applied after each layer. In the BCQ methods, the linear layer is common to both the Q-network and the clinician network, but the clinician network also includes an additional two linear layers, each with 128 nodes. The final output of the clinician network passes through a softmax activation function to determine the probabilities for each action.
The hyper-parameters used across all algorithms are uniform and detailed in Table \ref{tab:table2}.

\begin{table}[h]
\caption{Hyper-parameters of Networks}
\centering

\begin{tabular}{l|l}
\hline
Hyper-parameter                 & Value         \\ \hline
Network optimizer               & Adam          \\
Learning rate                   & 0.00005       \\
Adam (epsilon)                  & 0.0001        \\
Discount factor(gamma)          & 0.99          \\
Mini-batch size                 & 32            \\
Target network update frequency & 500 episodes  \\
Target network update method    & Polyak update \\
Threshold (tau)                 & 0.3           \\
Replay buffer size              & 5$\times10^4$             \\ \hline
\end{tabular}%
\label{tab:table2}
\end{table}

\subsection{Policy Evaluation}
To assess the effectiveness of RL-based treatment policies, we used the MIMIC-III dataset. We conducted policy learning using a state space comprising 16 features and a discrete action space with six categories. The dataset was split into an 80\% training set and a 20\% test set. Performance evaluation was based on the expected return calculated using Weighted Importance Sampling (WIS). 

\begin{figure*}[h]
    \centering
    \includegraphics[width=0.75\textwidth]{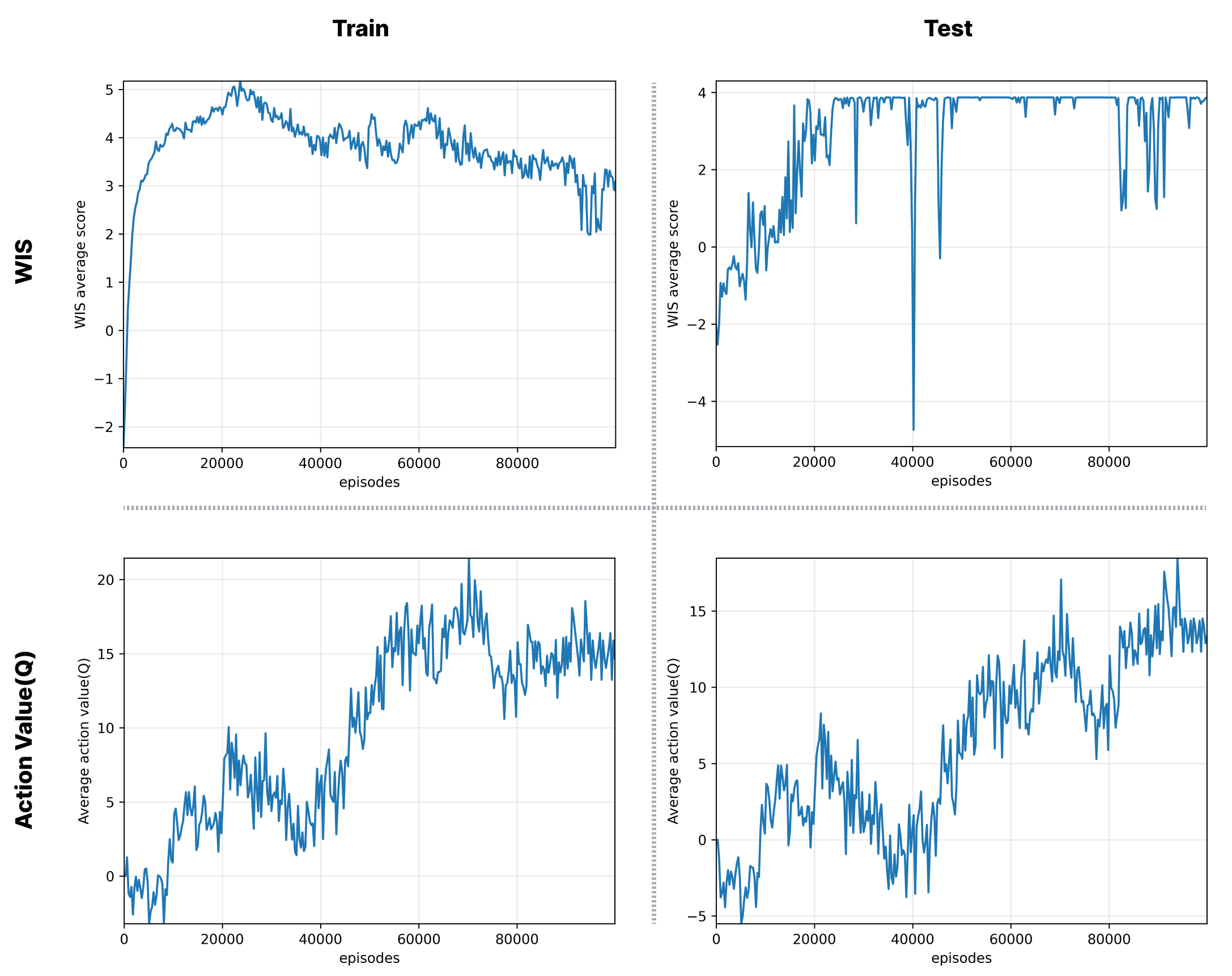}
    \caption{Training Curves for Average WIS and Predicted Q-Values}
    \label{fig: figure4}
\end{figure*}

Through analysis of actual ICU patient care data, we successfully trained the heparin dosing policy. The policy's development over time, indicated by WIS and average predicted Q-values, is depicted in Fig. \ref{fig: figure4}. The training involved 100,000 episodes utilizing an experience replay buffer containing the entire set of patient trajectories. The performance was assessed every 200 episodes, with each dot on the graph representing one evaluation. The graphs display the average off-policy evaluation performance per episode and the average predicted Q-values; the left side shows results from the training set, and the right side shows results from the test set. Over the course of training, both WIS and action values progressively increased, reflecting a positive reinforcement of the policy according to the defined reward function.

Furthermore, to assess our approach's effectiveness, we compared the BCQ with several value-based DRL algorithms known for their efficacy across various RL domains. Specifically, we evaluated deep Q-learning (DQN), dueling deep Q-learning (Dueling DQN), and double deep Q-learning (Double DQN). We also examined the performance of these RL policies against existing clinical policies to determine if they offer improvements over standard clinician-driven approaches.

\begin{figure*}[h]
    \centering
    \includegraphics[width=0.75\textwidth]{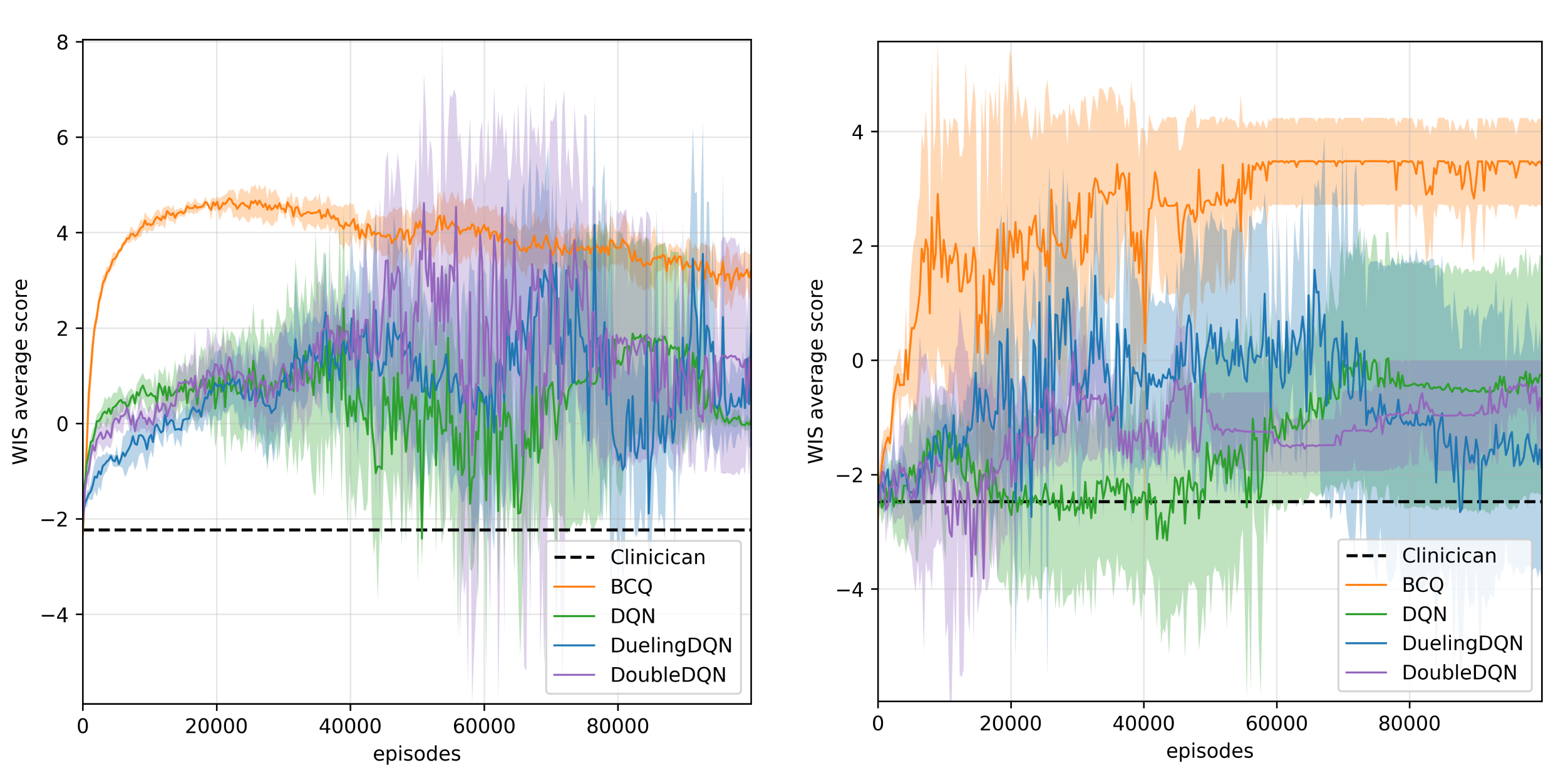}
    \caption{Comparative Performance of Off-Policy RL Algorithms}
    \label{fig: figure5}
\end{figure*}

Fig. \ref{fig: figure5} presents the results, showing the estimated cumulative rewards for each policy. The experimental setup mirrors that of Fig. \ref{fig: figure4}, with mean values and standard deviations visualized based on five random seeds. The left graph displays results from the training set, while the right graph shows outcomes from the test set. Notably, the expected return of BCQ demonstrates stable convergence within a specific range. Among the various algorithms, the policy trained using BCQ consistently outperforms others, exhibiting less variance and indicating that the Batch-Constrained approach contributes positively to consistent learning and enhanced performance in offline RL settings. Additionally, as the number of episodes increases, the RL-derived policy consistently surpasses the expected rewards of the clinician's policy, underscoring the effectiveness of the RL approach in improving heparin treatment.

To quantitatively compare the algorithms, we analyzed the expected returns using the highest-performing agent from the training episodes. Table \ref{tab:table3} presents these expected returns for the test set. The results reveal that all RL policies outperform the clinician's policy in terms of expected returns. Among them, the BCQ method exhibits the highest performance, with significant leads over DQN, Double DQN, and Dueling DQN.

\begin{table}[h]
\caption{Expected Returns for RL Algorithms and Clinician Policy}\label{tab:table3}
\begin{tabularx}{\linewidth}{@{}l*{5}{>{\centering\arraybackslash}X}@{}}
\toprule[1.5pt]
    & BCQ  & Double DQN & Dueling DQN & DQN  & Clinician \\ \midrule
WIS & 3.48 & 0.14       & 1.58        & 0.12 & -2.47     \\ \bottomrule[1.5pt]
\end{tabularx}
\end{table}

\begin{figure*}[t]
    \centering
    \includegraphics[width=1.0\linewidth]{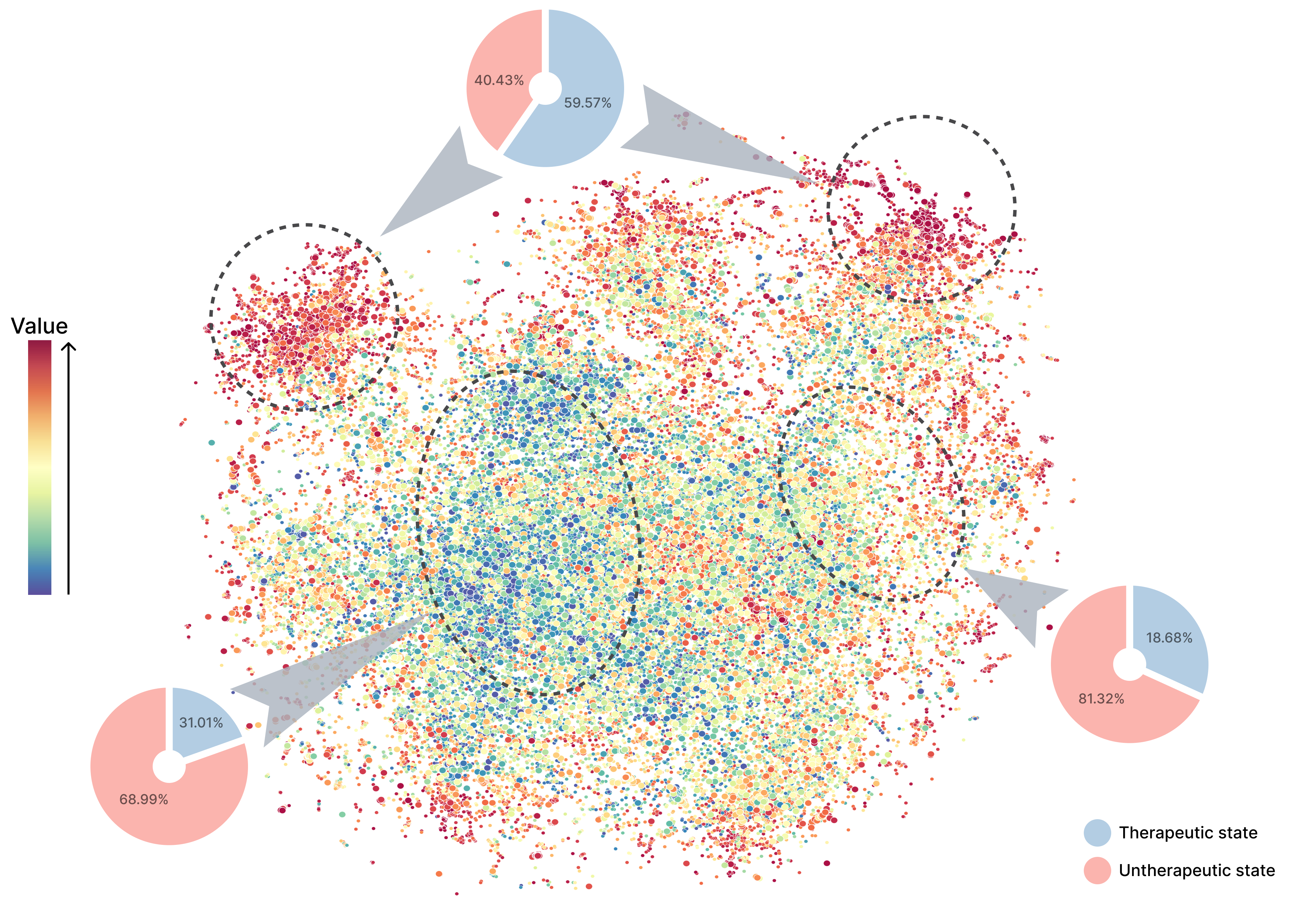}
    \caption{t-SNE Visualization of State Representations by Expected Reward}
    \label{fig: figure6}
\end{figure*}

\subsection{State Representation With Maximum Expected Reward}
We utilized t-SNE to visualize the state representations under the RL medication policy. 
Fig. \ref{fig: figure6} shows the results of this analysis, where the BCQ agent's activity against a replay buffer was evaluated, and t-SNE was applied to the state features. Each point in the figure is colored based on the maximum expected reward predicted by BCQ for that particular patient state, with colors ranging from red (highest value) to blue (lowest value). Larger points indicate data from the test set, while smaller points represent the training set.

To qualitatively analyze the states based on their value, we examined the percentage of therapeutic aPTT range for states classified into high, medium, and low-value regions. In the pie chart, `Therapeutic' states indicate the proportion of points where the aPTT values are within the optimal range of 60 to 100, and `Untherapeutic' states represent the proportions outside this range.

The analysis revealed that areas assigned high values (top right of the figure) contain a significant number of patient states corresponding to the `Therapeutic' state, suggesting that these states are either achieving immediate rewards or have already reached the optimal state. Conversely, regions assigned low values (middle of the figure) show a lower proportion of `Therapeutic' states, indicating that these patients are far from the desired stable state, thus far from achieving rewards. Medium value regions (top left) show intermediate proportions, aligning with expectations. The t-SNE algorithm effectively groups similar BCQ state representations close together, demonstrating that the policy learned by BCQ aligns well with the objectives defined by the reward function.

\subsection{Action Distribution}
\begin{figure}[h!]
    \centering
    \includegraphics[width=1.0\linewidth]{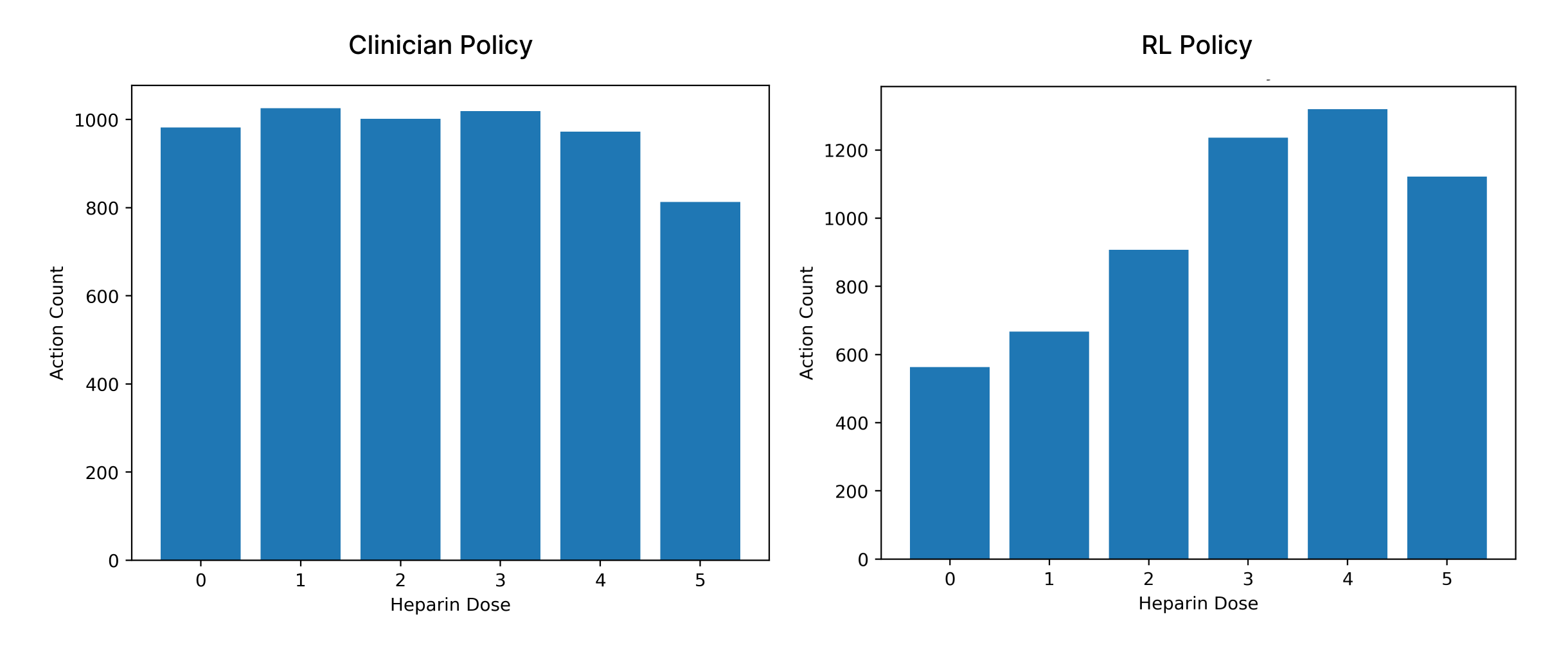}
    \caption{Comparison of Action Distributions Between Clinician and AI Policies}
    \label{fig: figure7}
\end{figure}

To gain a deeper understanding of the RL policy, we analyzed and compared the actions taken by the clinician policy and the RL policy. Fig. \ref{fig: figure7} displays the aggregate results of all actions selected by each policy at every timestep within the test dataset. The number of actions indicates how frequently a specific heparin dose is administered.
From the analysis, it is apparent that, in comparison to the clinician's policy, the AI's policy tends to administer higher doses of heparin more frequently. While the appropriate dosage is contingent upon the patient's condition, the AI policies developed in this study demonstrate a propensity for significant variability, particularly from the perspective of aPTT management.

\section{Conclusion}\label{sec5}
We adopted a batch-constrained approach to effectively learn RL policies using EMR data. As evidenced by Table \ref{tab:table3} and Fig. \ref{fig: figure5}, the performance of traditional deep reinforcement learning (DRL) algorithms in offline settings is often limited. By integrating an expert behavior network, we mitigated the overestimation of Q-values and achieved a balance between the behavioral policy and optimal policy estimation using the Bellman equation. The BCQ algorithm demonstrated superior performance over existing DRL approaches.

Additionally, we employed t-SNE to analyze the relationship between state representations and expected values, confirming that the policy accurately learns the objectives set by the reward function. High-value states were found to be closer to achieving the desired reward, whereas low-value states were distant. This fixed dataset training system without environmental interaction is particularly apt for the medical field, where simulations are subject to stringent constraints.

Decision support systems powered by artificial intelligence show promise in thrombosis treatment. Nonetheless, understanding the directional intent of DRL policies is crucial for their practical application. As shown in Fig. \ref{fig: figure7}, the RL policy tends to favor a more aggressive strategy to maximize rewards. While there is no universally correct approach to heparin dosing, it is vital for clinicians to comprehend the tendencies of DRL policies when applied in real-world scenarios. If needed, the policy’s aggressiveness can be moderated through strategic constraints.

This study has limitations, including the use of discrete actions and reliance solely on activated Partial Thromboplastin Time (aPTT) as the reward function. Given that drug dosing is typically a continuous variable, future improvements should include transforming the action space to accommodate more precise dose predictions. Moreover, we aim to refine the reward function to better reflect induction into the therapeutic range, changes in the patient’s organ function, and their clinical implications. Future research will focus on designing an enhanced RL framework to incorporate these aspects.

\bmhead{Acknowledgements}
This work is the result of the research funded by the National Research Foundation of Korea (NRF) under Grant 2021R1F1A1061093.

\section*{Declarations}
\begin{itemize}
\item \textbf{Funding}: This study was funded by the National Research Foundation of Korea (NRF) under Grant 2021R1F1A1061093.
\item \textbf{Competing interests}: The authors have no competing interests to declare that are relevant to the content of this article.
\item \textbf{Ethics approval}: Not applicable.
\item \textbf{Consent for publication}: All authors consent to the publication of this manuscript.
\item \textbf{Data availability}: The data used in this study are openly available in the MIMIC-III database (https://mimic.physionet.org/).
\item \textbf{Materials and Code availability}: The materials and code used in this study are available upon request.
\item \textbf{Author contribution}: Conceptualization: Inbeom Park, Sujee Lee; Methodology: Yooseok Lim, Inbeom Park, Sujee Lee; Formal analysis and investigation: Yookseok Lim; Writing - original draft preparation: Yooseok Lim; Writing - review and editing: Inbeom Park, Sujee Lee; Funding acquisition: Sujee Lee; Resources: Sujee Lee; Supervision: Sujee Lee.

\end{itemize}

\noindent

\bibliographystyle{sn-basic}
\bibliography{sn-article}

\end{document}